\documentclass{sigchi}

\CopyrightYear{2020}
\setcopyright{acmlicensed}
\doi{https://doi.org/10.1145/3313831.XXXXXXX}
\isbn{978-1-4503-6708-0/20/04}
\conferenceinfo{CHI'20,}{April  25--30, 2020, Honolulu, HI, USA}
\acmPrice{\$15.00}
\usepackage{balance}       
\usepackage{graphics}      
\usepackage[T1]{fontenc}   
\usepackage{txfonts}
\usepackage{mathptmx}
\usepackage[pdflang={en-US},pdftex]{hyperref}
\usepackage{color}
\usepackage{booktabs}
\usepackage{textcomp}
\usepackage{microtype}        
\usepackage{ccicons}          
\usepackage{todonotes}
\def\plaintitle{SIGCHI Conference Proceedings Format}

\def\emptyauthor{}
\def\plainkeywords{Authors' choice; of terms; separated; by
  semicolons; include commas, within terms only; this section is required.}

\makeatletter
\def\url@leostyle{%
  \@ifundefined{selectfont}{
    \def\UrlFont{\sf}
  }{
    \def\UrlFont{\small\bf\ttfamily}
  }}
\makeatother
\def\pprw{8.5in}
\def\pprh{11in}

\definecolor{linkColor}{RGB}{6,125,233}
\hypersetup{%
  pdftitle={\plaintitle},
  pdfauthor={\emptyauthor},
  pdfkeywords={\plainkeywords},
  pdfdisplaydoctitle=true, 
  bookmarksnumbered,
  pdfstartview={FitH},
  colorlinks,
  citecolor=black,
  filecolor=black,
  linkcolor=black,
  urlcolor=linkColor,
  breaklinks=true,
  hypertexnames=false
}

\begin{document}

\title{Building domain specific lexicon based on TikTok comment dataset}
\numberofauthors{3}
\author{%
  \alignauthor{Jiaxiang Hao\\
    \affaddr{School of Computer Science, The University of Adelaide}\\
    \email{jiaxiang.hao@student.adelaide.edu.au}}\\
}
\maketitle

\begin{abstract}
   In the sentiment analysis task, predicting the sentiment tendency of a sentence is an important branch. Previous research focused more on sentiment analysis in English, for example, analyzing the sentiment tendency of sentences based Valence, Arousal, Dominance of sentences. the emotional tendency are different between two languages. For example, the sentence's order between Chinese and English may present different emotions. This 
  paper tried a method that build a domain specific lexicon. by this way, model can classify Chinese words by emotional tendency. In this approach, based on the \cite{DBLP:conf/naacl/RotheES16}, an ultra dense space embedding table be trained through  word embedding of Chinese TikTok review and emotional lexicon sources(seed words). The result of model is a domain specific lexicon, which present the emotional tendency of words. I collected Chinese TikTok comments as training data. By comparing The training results with the PCA method to evaluate the performance of the model in  Chinese sentiment classification, the results show that the model done well in Chinese. The source code has released on github: \url{https://github.com/h2222/douyin_comment_dataset}
\end{abstract}


\section{1. Introduction}
Deep learning is developing rapidly in many areas of Natural Language Understanding. While the deep learning requires a large amount of data to ensure the generalization of the model. For example, in the famous pre-trained language model ''BERT'' \cite{devlin-etal-2019-bert} where Google provides 400 million data to train Bert model, which requires a huge amount of data to guarantee the model performance. However, in many cases we cannot obtain a large amount of data. For example, training data is scarce in the task of word emotional classification.\\
Previous research \cite{DBLP:conf/acl/HatzivassiloglouM97} tried to get the "semantic orientation" of the sentence. For example, positive and negative tendencies. through obtaining the predictions of emotion tendencies and comparing the predictions with the real sentences, the accuracy of the prediction is obtained.\\
Further, the researchers found that, based on psychological research, multiple factors will lead to different emotions rather than solely determined by emotional polarity. For example, in the study of Bradley and Lang's research \cite{Measuring-emotion}, Valence-Arousal-Dominance model to determine the emotional tendency from three different factors.\\
Word embedding is an effective way to represent language features, such as similarity\cite{DBLP:conf/emnlp/PenningtonSM14}, emotional orientation \cite{DBLP:conf/naacl/RotheES16}, and sentiment\cite{DBLP:journals/ci/CalvoK13}. Word embedding is the method to victories strings, different dimensions of vectors may contain different information, but embedding representation is "black box", it is hard to know what exactly mean the dimension present.\\
In this paper, we proposed that emotional information can be compressed as a one dimension embedding called ultradense subspace, we hypotheses the ultradence subspace as the space where the sentiment be quantified as numerical values and be classified by trained ultradense word embedding. For Chinese and English, we predict that the emotional words of two languages can correctly classify but not restricted by language types and grammars.\\
The advantage of this method is that even if the data is lacking, the model has good performance. And the data does not need to be labeled, so there is a large amount of data that can be used to train. Another advantage is that, compared to the origin word embedding, the ultradense subspace has higher performance and quality. For example, in classification tasks, unrelated information is considered as noise, which is filtered through orthogonal transformation. As a result, the model has fewer parameters, which means that the training speed is improved and the possibility of overfitting is reduced.\\
In this paper, our goal is to produce a decent word embedding in Chinese. The output of model will be a lexicon, which is converted from the original embedding, the embedding of lexicon presented as one-dimensional ultradense embedding table, this table produce a clearer, easier-to-explain word representation for word emotions.

\section{2. motivation}
In most of the previous studies, the models were trained using English datasets, and rarely use Chinese datasets, especially web review data, the web review contain stronger emotional tendencies than the written context. I would like to do Chinese emotional classification by using this method, which transforms embedding into a low-dimensional, interpretable embedding. Based on the output of the embedding, we can know whether the model is well in Chinese emotion classification. At the same time, I am also looking for factors that decrease the performance of the model. For example, according to research \cite{DBLP:conf/naacl/RotheES16}, they find that some factors of embedding such as embedding size, stop words, embedding dimensions, and seed words may effect performance of model. Thus, we decide to evaluate what kind of factors 
influence the mode results.\\

\section{3. Background}
In this section, I will introduce some relevant background about emotional expression and word embedding, and some of techniques will use in our methodological work. 
\subsection{3.1 emotion representation}
Emotional expression is not only distinguished by positive and negative, but also by multiple factors. The VAD model (Figure 1) \cite{Measuring-emotion} considers that emotions to be composed of three different dimensions, named Valence (positive-negative), Arousal (clam-excited), and Dominance (perceived degree of control situation).  There are also studies \cite{DBLP:conf/hci/ZhongQZ19} which suggest that Dominance does not affect the expression of emotion (only VA model works). As the most common system for assessing emotional tendencies, the VAD model quantifies Valence, Arousal and Dominance, and then evaluates them.
\begin{figure}
\centering
  \includegraphics[scale=0.6]{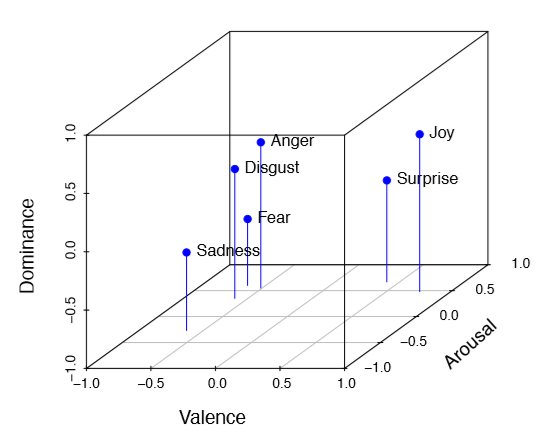}
  \caption{The emotion space covered by the Valence-Arousal-Dominance (VAD) model}~\label{fig:figure1}
\end{figure}

\subsection{3.2 emotion lexicon}
In some psychological studies, words with emotional tendencies illustrated as specific numerical values. For example, if the value of a word is greater than a certain threshold, the word is considered to be a positive emotional word, whereas if a word is less than the threshold, it is Negative emotion word. In this paper, the emotion lexicon that we build only use positive and negative emotions as a one-dimensional evaluation. For illustration, we will visualize the result as a 2d figures to build the lexicon that the emotional values of word distributed on the figures.\\

\subsection{3.3 word embedding}
Word embedding is a vector representation, which encodes word into a vector form. In this way, word is converted to the numeric values, it can use into unsupervised learning. Here are some very popular embedding algorithms today: WORD2VEC is a popular embedding algorithm, which provides a trimmed down neural network \cite{DBLP:conf/nips/MikolovSCCD13}. FASTTEXT is based on WORD2VEC, which is a derivative of WORD2VEC and combining the n-gram of characters \cite{DBLP:journals/tacl/BojanowskiGJM17}. Unlike the previous embedding algorithms, GLOVE directly trains word vectors on the word co-occurrence matrix, which improves the training efficiency.

\subsection{3.4 word-level prediction}
In previous research \cite{DBLP:journals/tois/TurneyL03}, researchers distinguished the emotional tendency of words by using the polarity of words. The specific approach is to divide words into two kind of training data, the original word embedding and the seed words. The seed words are usually positive or negative The word, the distance between the pair of seed words is obtained point-wise, so as to obtain the emotional tendencies of words. Afterwards, Rothe \cite{DBLP:conf/naacl/RotheES16} proposed to transform the word embedding from the original space to ultradense subspace by training the orthogonal matrix, which will get embedding of word's emotion.\\

\section{4. Research Method}
In order to analyze the sentiment of TikTok comment data and build TikTok sentiment lexicon, I need to collect and process reviews from the TikTok app. My plan is to use web crawlers to crawl data from the TikTok app and preprocess and victories the data,  which need to implement a web crawler framework to collect data for the TikTok app, then use python data analysis tool Pandas to pre-process review data and use WORD2VEC training tool Gensim to victories review sentences, the processed word embedding table saved as the VEC files. For seed words, I labeled 5, 10, 15 for seed word groups. In the end, After trained model, I visualized and analysed results.\\

\section{5. Research Questions}
First, "Is there a relationship between the seed words and the emotions?" When the size of word reaches a certain amount, the generalization ability of the model improved, that is, the model may learn more prior information. While the model learned the distance between seed words, we can verify that whether the distinguishing ability of model be affected when the size of seed words change. One hypothesis is that the model's ability to distinguish emotions grows if we expands the size of the seed word. To verify the hypothesis, I set the size of seeds to 5, 10, 15 respectively, and each seed word group join the training processes. Finally, 3 results will be visualized to verify the hypothesis.\\
The second question is "Does a specific video guide the commentator's emotions?" When we collected data, we found that the commentator's emotional inclination would be consistent with the emotional tendencies that the video wanted to express. For example, most Comments are sad in videos of funerals, but most of the comments are pleasing in funny videos. Therefore, we made a hypothesis that the sentiment of a particular video will guide the comments emotional tendencies. To verify the hypothesis, we collected two different data, comment data from random videos and comment data from special videos, I use different datasets to train the model, and finally compare the distribution of words.\\
The third question is "Compared with the previous method, is there any improvement in compressing embedding to ultra dense subspace?" In the previous research, the commonly used approach of feature extraction is Principle Compose Analysis, PCA is the feature selection method, which is a commonly used tool in data analysis and model prediction. It is used to visualize the distance and relationship between data. PCA uses Singular Value Decomposition to select appropriate dimensions to achieve the goal of dimensionality reduction of data. In this paper, after processing data by PCA and the ultradense compression, we will compare the results to find that which method can better show the emotion of words. One hypothesis is than the data processed by ultradense compression has the better performance on emotional classification task, because ultradense compression uses seed words, which reduces the distance between the same emotion words and increases the distance between different emotion words. To verify this hypothesis, we will use the same training data to get results from PCA and ultradense compression, and then we will visualize and analyze the results to verify hypothesis.

\section{6. Data Collection}
\subsection{6.1 collection}
The data we need to collect are user's comments from TikTok random videos and user's comments from particular videos. The comment data from random videos can analyse the user's sentiment for making sentiment lexicon, and the comment data of specific videos can analyze the user's sentiment expression under specific videos. People's sentiment often be reflected in words, and words with sentiment appeal are often commented by people. As a result, after the vectorization of comment data in vector space, the vector that have same sentiment are often collected together. Through research on RQ1 We can get the distribution of the vector of random video comments in vector space, which will tell us what sentiments expression will show what vector distribution.\\
We randomly crawl TikTok comments data through the web crawler, and crawl specific comment data under some very high like videos. These comment data can be found at \url{https://github.com/h2222/douyin_comment_dataset}
the data contains 3 type of files. The dataset.csv file represents the original data crawled from TikTok where data preprocessing is required. The fixdata.py file is a data processing file for processing raw data. The processed file saved in dataupdate.xlsx file For further use, the data collects the user's age, gender, nickname, comments, number of comments liked, etc.

\subsection{6.2 preprocess}
The data will be saved as a csv file, and a new repository will be created to save the data on GitHub. The data set to two categories: random video comment data and specific video comment data. For the dataset, the related heads includes the user's nickname, age, gender, number of likes, and comment details, where gender head can be binarized, such as 0 for female and 1 for male. In addition, comments can be grouped with users of different ages to research if age affects sentiment expression. The Python Pandas module use to analyse and preprocess the data and Gensim module use to train a original embedding table by pre-processed comment data,  we will also labeled emotional words as seed words.

\section{7. Data Analysis}
First of all, I want to evaluate the performance of the model When using a different number of seed words. I decided to plot the distribution of a word in the case of using different seed words. the figure 2 is the Chinese TikTok words distribution, which is based on the training results of the model, the  emotional distribution can be shown as the distance between words.\\
\begin{figure}
\centering
  \includegraphics[scale=0.25]{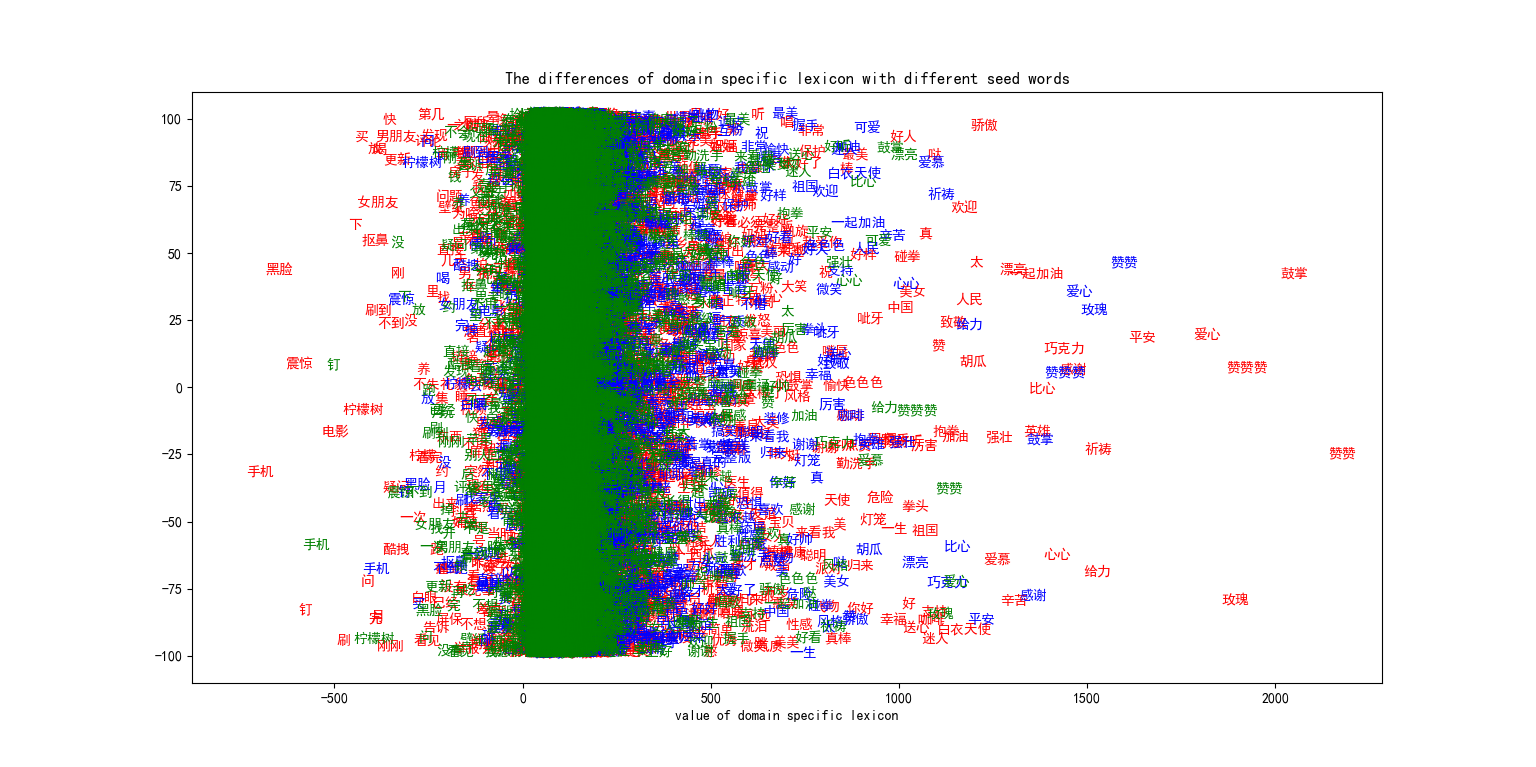}
  \caption{the words distribution based on different seed words, red part is 5 seeds result, blue part is 10 seeds result, green part is 15 seeds result }~\label{fig:figure1}
\end{figure}
According to the word distribution, the X axis is the number of all words, which comes from the TikTok comment data after the word segmentation. The Y axis is a random number between 100 and -100, the reason we add the Y axis is that the word distribution is 1D distribution, which means that the experiment only obtains positive and negative emotions, and the distribution results will be "crowded" on a line. In order to better visualize the distribution results, we add a random number as the second dimension, so The words are mapped on the 2D plane.\\
Looking at the image, we find that the model's ability to distinguish positive emotions is stronger than that of negative emotions. For example, if we think that the position where index equals 0 is the distribution of neutral words, the minimum value of negative words are close to -600. Instead, The maximum value of positive words are close to +2000. Evaluating from the semantics, the model are also better distinguishes positive sentiment words. I think the reason for that is because there are more positive words in the TikTok dataset than negative words, so it will Let the model produce better training results to positive words.\\
Further, we found that the size of the seed words may affects the ability of emotional classification. For example, under the condition of training the model with same dataset, we found that when the seed word is 15 (green distribution in the figure), the positive word The maximum value is +1000 and the minimum value of the negative word is -250, and the span is approximately 1250. When the seed word is 5 (the red part in the figure), the maximum value of the word is +2000 and the minimum value is -600, and the span is 2600. Therefore , the smaller the seed words, the stronger the model's ability to distinguish sentiment words. In my opinion, during the training process, the model distinguishes word sentiment by optimizing the distance between seed words and embedding words. Excessive addition of seed words may increase the model's Parameters, which leads to overfitting of the model. Based on the experiments, when the size of the seed words equal to 5, the model's result is the best.\\
\begin{figure}
\centering
  \includegraphics[scale=0.25]{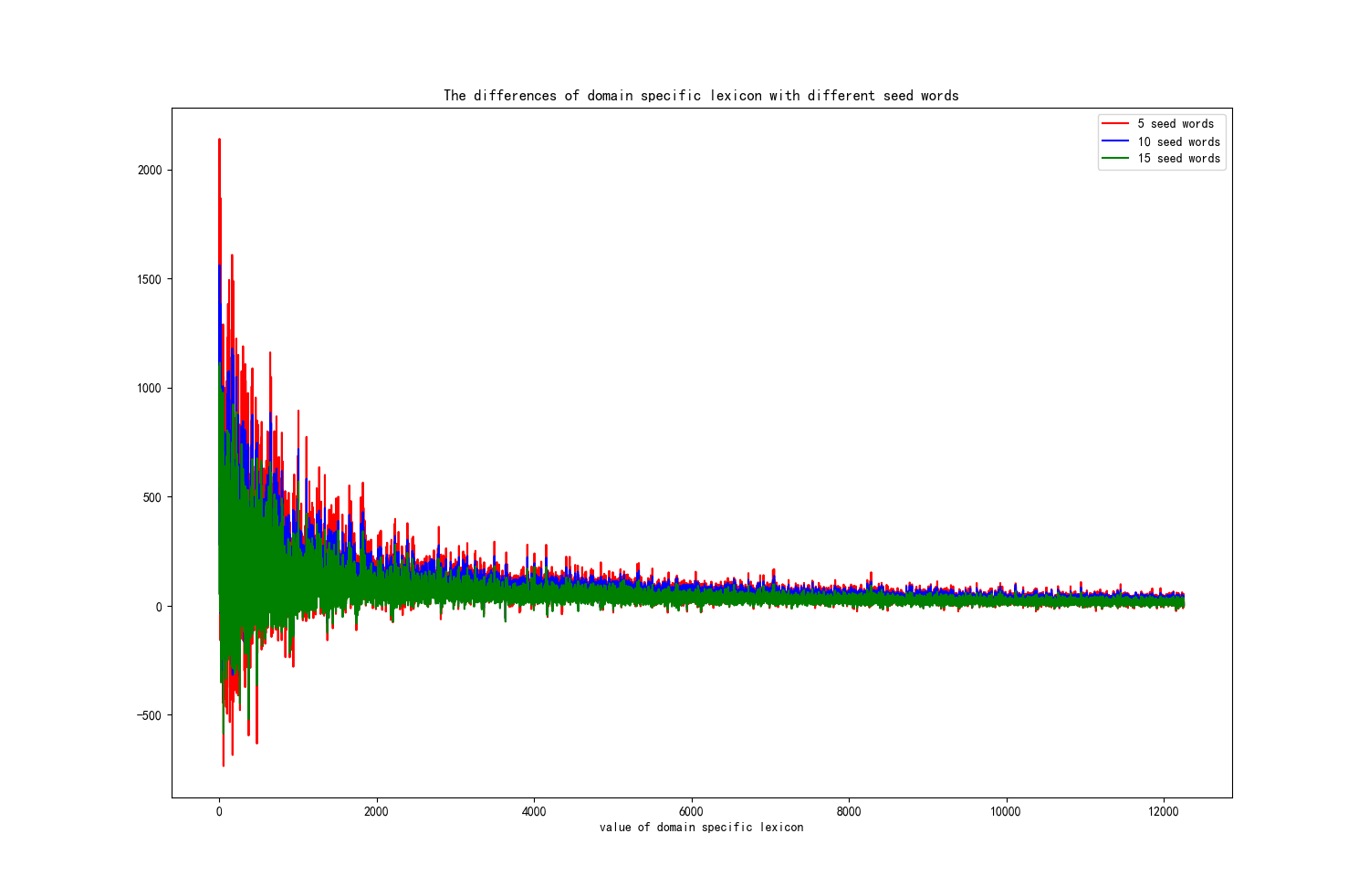}
  \caption{the result embedding value against the set of words, red line is 5 seed result, blue line is 10 seed result, green line is 15 seed result }~\label{fig:figure1}
\end{figure}
Next, we draw a line graph to show the model's distinction between sentiment words. As shown in figure.3, the X axis is the number of all words, and the Y axis represents the output value of each words. According to the figure, we found that the model can be more distinguish the first 2000 words to a large extent of which 1 to 2000 words have a larger interval on the Y axis. When the word is after 2000, the model gradually loses its ability to distinguish the emotion of the words.\\
Next, we sort the results. The model converts words embedding table (shape is [vocabulary size, embedding size]) in the original space to words representation (shape is [vocabulary size, 1]) in ultradense subspace. Therefore, we sort all the words according to the values in the new 1D space. The result of sorting is shown in figure 4.\\
\begin{figure}
\centering
  \includegraphics[scale=0.25]{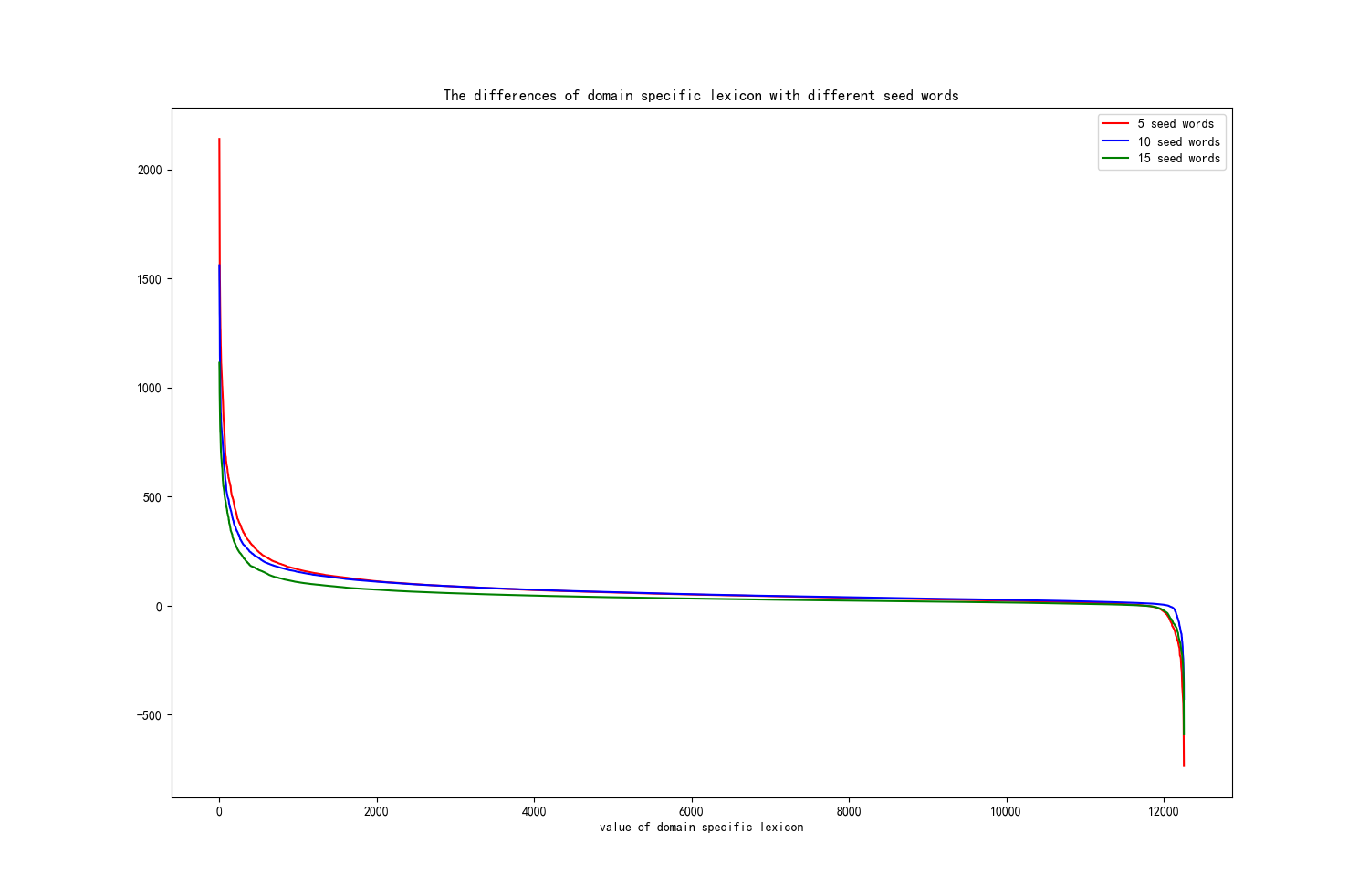}
  \caption{the line graph of sorted words, red line is 5 seed result, blue line is 10 seed result, green line is 15 seed result }~\label{fig:figure1}
\end{figure}
We found two obvious upward and downward trends on both sides of Figure 4, which shows that the word representation can distinguish the sentiment of the word in the ultradense subspace. When the seed words are 15 (the green part in the figure), the turning point is more obvious, I guess the increase of seed words may increase the model's ability to classify fuzzy words. For example, some nouns words with emotional tendencies such as 'cake', 'firework' can be correctly classified.\\
\begin{figure}
\centering
  \includegraphics[scale=0.25]{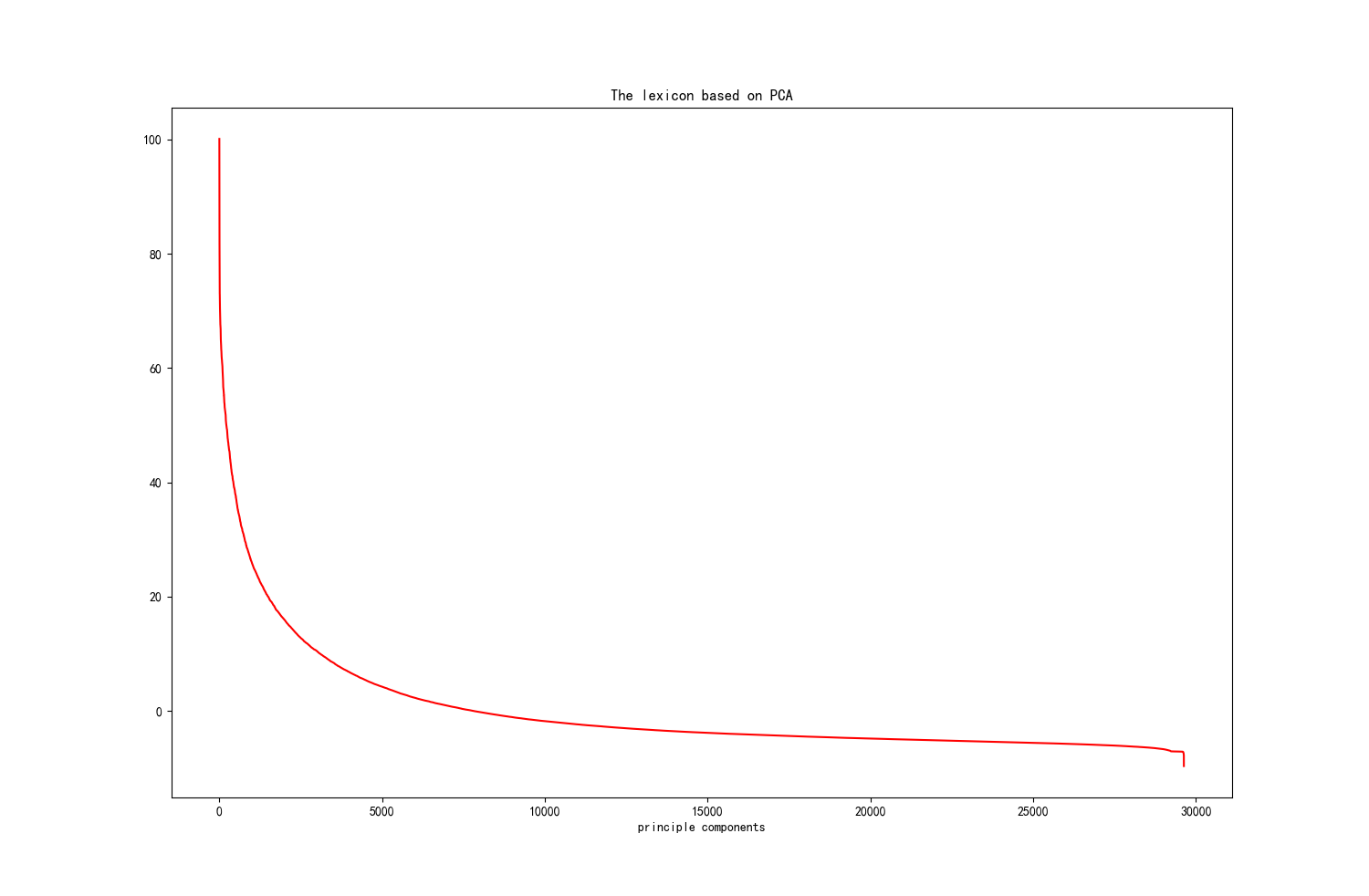}
  \caption{the result of PCA,  the X axis is words index sorted by embedding value of PCA, the Y axis is the word embedding of PCA}~\label{fig:figure1}
\end{figure}
In order to prove that the emotion can be better represented by using the ultradense compression method, we use PCA for comparative experiments. The original word embedding table (shape is [vocabulary size, embedding size]) is compressed into a new embedding ( shape is [vocabulary size, 1], shape is consistent with that of result of ultradense compression), as shown in Figure 5.\\
As shown in Figure 5, the index of the sorted words on the X axis is based on the word embedding value of PCA, and the Y axis is the word embedding value. Compared with the orthogonal transformation method (Figure 4), the curve of the PCA-based method is more gentle, This shows that PCA cannot distinguish the emotion of words well. Moreover, the PCA method has insufficient predictive ability for negative emotion words.

\section{8. Findings and discussion}
\subsection{8.1 Is there a relationship between the seed words and the emotion?}
A result shows that there is indeed a relationship between seed words and emotional tendencies, but the word emotion may more related with training data. An obvious conclusion is that most of the comments use positive words but negative words have not been widely appeared in the comments, which can be verified by figure 6, The interesting result is that words with a large ultradense embedding value appear very frequently in comments. These words usually have a positive meaning. table 1 are some of the most frequent words and the number of occurrences.\\
\begin{table}
  \centering
  \begin{tabular}{l r r r}
    {\small\textit{word}} & {\small \textit{numbers}}\\
    \midrule
    feel & 3710  \\
    Awesome & 3523  \\
    heart & 3365\\
    praise & 2351  \\
    ...\\
  \end{tabular}
  \caption{Table captions should be placed below the table. We
    recommend table lines be 1 point, 25\% black. Minimize use of
    table grid lines.}~\label{tab:table1}
\end{table}
In the comment dataset, the most of high frequent words have positive meanings, even when we set the seed words, we found that 30\% of the seed word appeared in the top 200 words of comment dataset. Combined with the previous model result, we guess that if words with certain type appear frequently, the word is easier to be distinguished by the model. In addition, we found that in the dataset, in the comments that contain high-frequency words, the comment with 2 or more The high-frequency words are 44\%. In other words, nearly half of the high-frequency words appear in pairs in the comments, which will cause the associated relationship when the word is converted into embedding. As the result, the model is affected.\\

\subsection{8.2 Does a specific video guide the commentator's emotions?}
Special model videos will impact on the comments. the five different categories of video comment are collected and the categories are health, star, funny, news and entertainment. Through model training to get emotional lexicon, we found that the emotional tendency is biased to positive in funny videos, but the emotional tendencies of comments are more neutral in news videos. \\
I guess the reason for this result is that the content of the video will guide the emotional tendency of the comment, so that the frequency of the words that match the emotional tendency of the video will increase, and the model can better learn the emotional information of those words, thereby improving the model The ability to discern specific emotions. For example, when processing the comment data of funny videos, we found that some words with a clear positive tendency have a high Term Frequency (other words divided by the most frequent words), such as Awesome, funny , happy face, the Term Frequency of these words can be close to 0.9. Finally, we use the model to classify the emotions of the words, and find that the model has a more obvious effect on the classification of words with high frequency.

\subsection{8.3 Compared with the previous method, is there any improvement in compressing embedding to ultra dense subspace?}
After testing, we found that the performance of ultra dense word embedding is better than that of PCA method in sentiment classification tasks. Comparing Figure 5 and Figure 6, we observed a clear distinction between words using the ultra dense embedding method (the turning point in Figure 5). On the contrary, although the PCA method can distinguish words (the curve in Figure 6), it cannot clearly distinguish words, and the curve is relatively smooth, so PCA cannot distinguish the emotion of words well.\\
I think that ultra dense word embedding performs better because the model learns emotional information from seed words. The model reduces the distance of the same emotional words in the seed word and increases the distance of the opposite emotional words, so that the transformed representation of the model learns the emotion Information, which improves the word emotional representation. The PCA method does not analyze the emotion of the word, so ultra dense word embedding is better in the emotion classification.\\

\section{9 Threats to Validity}
My paper may still have some disadvantages. First, my data collection volume is small and limited. Since English review data cannot be collected, we only build sentiment dictionaries on Chinese data, because both Chinese and English are different in grammar and expression, I expect differences in different language data sets, so a comparative test is necessary To evaluate the performance difference of the model when dealing with different languages. \\
secondly, the amount of data we collected did not meet expectations. We spent a long time on collecting TikTok's review data. As a very popular application, TikTok's company designed a complex encryption algorithm to protect user privacy, This is a very good thing, but for researchers who want to perform data analysis, they have to spend a lot of time users cracking the anti-crawl mechanism of TikTok. When collecting data with comments, it took us a long time to crack TikTok ’s x-gergon encryption algorithm, which is a hexadecimal encryption based on MD5, and it is difficult to crack. As a result, the data I collected did not reach the expected amount. In the future research, we will strengthen the data collect.\\
Finally, I think that in the experimental stage, the performance of the model can be further optimized. In the training process of the model, because we need to reduce the distance between the same emotion words and increase the distance between different emotion words, we need to optimize the same and different emotions at the same time. The loss of the word. Therefore, we use a loss function with hyper-parameter alpha, which can combine the two losses in a certain proportion, so that the global loss can be optimized. The formula is as follows:\\
\begin{align*}
    Loss = (1 - \alpha) \times SLoss + \alpha \times DLoss
\end{align*}
where $\alpha$ is the hyperparameter of the model, SLoss is the loss of the same sentiment words, and DLoss is the loss of different sentiment words.\\
During training, we found that when the iteration of the model reaches the 4th epoch (we set the model to iterate 10epoch), the loss of the model hardly drops. Regarding the cause of this problem, we guess that it is caused by the bias of the model data. In find and In the discussion section, we mentioned that positive emotion words have a higher frequency and higher in the data set, which leads to uneven distribution of positive emotion words and negative emotion words in the data set. Therefore, in the training stage of the model, 
the positive words are better optimized by the model compared to negative words, so the classification ability of the model for positive words is better than the classification ability of negative words.

\section{10 Related Work}

Faruqui \cite{DBLP:conf/acl/FaruquiD15} use word post-processing based on word similarity, so Faruqui does not use processing based on orthogonal transformation and does not need to consider word distance, which makes it worse for other applications such as syntax detection.The method of Faruqui does not benefit that ultra-dense embedding is more efficient.\\
In the tensor framework, Rothe and SchUtz \cite{DBLP:conf/acl/RotheS15} converts word embedding to perceptual (synonym) embedding. In their work, all the embeddings exist in the same subspace. However, we want to keep only specific information embedded in the subspace to create ultra-dense subspaces with a specific role (sentiment embedding).\\

The method we used is also related to directed PCA \cite{PCA}. However, compared to PCA, the directional PCA solution is not orthogonal.\\
The creation of the sentiment lexicon usually needs to label the training corpus. For example, people manually label the sentiment tendency of the words in the training set or form words with the same meaning into the training set, or add words with similar meanings to the seed set (\cite{DBLP:conf/acl/Turney02} and  \cite{DBLP:journals/jair/KiritchenkoZM14}). Heerschop team \cite{DBLP:conf/bis/HeerschopHF11} used wordNet and pageRank-based algorithms to propagate the emotional information of seed words into unknown words. Scheible team \cite{DBLP:conf/acl/Scheible10} proposed a semi-automatic method for machine translation Based on sentiment lexicon. Hamdan team \cite{DBLP:conf/semeval/HamdanBB15} evaluated the sentiment level based on the average of six sentiment lexicons. Those method cannot apply less-resource languages. Our experiments show that the orthogonal transformation method can train the sentiment lexicon with a small amount of training resources, and has better performance than the lexicon created by other semi-automatic methods.\\

\section{11 Conclusions}
We use the web crawler method to collect TikTok's Chinese comment data, and apply the method that can embed the original vector of data into the ultra-dense subspace. In this study, two experiments were used to verify that the ultra-dense subspace can classify the emotion of words. 1. By setting the number of different seed words, we get that the model can distinguish emotions through model training. 2. By comparing with the PCA method, ultra-dense word embedding can better distinguish the emotion of words. we only need to use 5-15 seed words as a training example to learn the subspace, and get the emotional information about the words through the subspace.

\section{12 Future Work}

In future work, we will find the probabilities that other information embedded in one or more orthogonal subspaces rather than just emotional embedding. This decomposition will not change the content of the information embedding, but it will make them more compact, meaningful and easier to interpret for any given application. In addition, we will use the larger dataset for training, because the size of the data determines the improvement of the model's generalization ability.

\bibliographystyle{plain}
\bibliography{ref.bib}

\end{document}